\documentclass[twoside]{article}
\usepackage[accepted]{aistats2016}

\usepackage{amsmath}
\usepackage{amsfonts}
\usepackage{bbm}
\usepackage{hyphenat}
\usepackage{natbib}
\usepackage[a4paper, total={7in, 8.75in}]{geometry}
\usepackage{authblk}
\usepackage{graphicx}

\makeatletter
\renewcommand\AB@affilsepx{, \protect\Affilfont}
\makeatother

\newcommand\indexSet{X}
\newcommand\subsetIndexSet{S}
\newcommand\dataSet{D}
\newcommand\inducingPointIdentities{Z}
\newcommand\augmentedSet{I}
\newcommand\projectionSet{C}
\newcommand\overallSet{\indexSet \cup \augmentedSet }
\newcommand\augmentationParameters{\theta}
\newcommand\indexSetMember{x}
\newcommand\holdOutSet{*}
\newcommand\genericIndexSet{U}
\newcommand\genericIndexSetB{V}

\newcommand\setOfAllCylinderSets{\mathcal{C}}
\newcommand\borelSigmaAlgebra{\mathcal{B}}

\newcommand\genericFunction{f}
\newcommand\indFunction{\genericFunction_{\inducingPointIdentities}}
\newcommand\dataFunction{\genericFunction_{\dataSet}}
\newcommand\dataFunctionNoInd{\genericFunction_{\dataSet \backslash \inducingPointIdentities} }
\newcommand\holdOutFunction{\genericFunction_{\holdOutSet} }
\newcommand\indexSetFunction{\genericFunction_\indexSet}
\newcommand\augmentedSetFunction{\genericFunction_\augmentedSet}
\newcommand\overallSetFunction{\genericFunction_{\overallSet}}
\newcommand\indexSetMemberFunction{\genericFunction_{\indexSetMember}}

\newcommand{\outputDataPoint}{y}
\newcommand{\outputDataSet}{Y}

\newcommand\nDataPoints{N}
\newcommand\nInducingPoints{M}
\newcommand\realNumbers{\mathbb{R}}
\newcommand{\realNumbersOverIndexSet}{\realNumbers^{\indexSet}}
\newcommand{\realNumbersOverDataSet}{\realNumbers^{\dataSet}}
\newcommand{\realNumbersOverInducingSet}{\realNumbers^{\inducingPointIdentities}}
\newcommand{\realNumbersOverAugmentedSet}{\realNumbers^{\augmentedSet}}

\newcommand\expectation{\mathbb{E}}
\newcommand{\KL}{\operatorname{\mathcal{KL}}}

\newcommand{\dee}{d}
\newcommand{\bigO}{\mathcal{O}}

\newcommand{\genericSet}{\Omega}
\newcommand{\genericSigmaAlgebra}{\Sigma}
\newcommand{\genericSubset}{E}
\newcommand{\genericSubsetB}{C}
\newcommand{\genericGeneratingSet}{G}
\newcommand{\genericMeasure}{\mu}
\newcommand{\candidateMeasure}{\hat{\genericMeasure}}
\newcommand{\genericMeasureB}{\eta}
\newcommand{\genericMeasureC}{\lambda}
\newcommand{\genericFunctionB}{g}
\newcommand{\genericSimpleFunction}{\psi}
\newcommand{\genericElement}{\omega}
\newcommand{\genericRangeValue}{y}
\newcommand{\genericRangeSetSize}{C}
\newcommand{\bayesRadonNikodymDerivative}{\frac{\dee \posteriorMeasure}{\dee \priorMeasure } }
\newcommand{\bayesRadonNikodymDerivativeMarginal}{\frac{\dee \posteriorMeasure_\dataSet}{\dee \priorMeasure_\dataSet } }
\newcommand{\approximatingRadonNikodyDerivative}{\frac{\dee \approximatingMeasure}{\dee \priorMeasure } }
\newcommand{\approximatingRadonNikodyDerivativeOverallSet}{\frac{\dee \approximatingMeasureOverallSet}{\dee \priorMeasureOverallSet} }
\newcommand{\approximatingRadonNikodyDerivativeMarginal}{\frac{\dee \approximatingMeasure_\inducingPointIdentities}{\dee \priorMeasure_\inducingPointIdentities } }
\newcommand{\approximatingRadonNikodyDerivativeMarginalAugmented}{\frac{\dee \approximatingMeasure_\augmentedSet}{\dee \priorMeasure_\augmentedSet } }
\newcommand{\transformationFunction}{h}

\newcommand{\genericRadonNikodymDerivative}{\frac{\dee \genericMeasure}{\dee \genericMeasureB}}
\newcommand{\projectionMap}{\pi}

\newcommand*{\LargerCdot}{\raisebox{-0.25ex}{\scalebox{1.8}{$\cdot$}}}

\newcommand\approximatingMeasure{Q}
\newcommand\approximatingMeasureOverallSet{\approximatingMeasure_{\overallSet}}
\newcommand\approximatingDensity{q}
\newcommand\priorMeasure{P}
\newcommand\priorMeasureOverallSet{\priorMeasure_{\overallSet}}
\newcommand\posteriorMeasure{\hat{P}}
\newcommand\posteriorMeasureOverallSet{\hat{P}_{\overallSet}}

\newcommand\specificRadonNikodymDerivative{\frac{\dee \approximatingMeasure }{\dee \posteriorMeasure}}
\newcommand\lebesgueMeasure{m}

\newcommand{\measurableSet}{A}
\newcommand{\measurableIndexSet}{\measurableSet_\indexSet}
\newcommand{\measurableInducing}{\measurableSet_\augmentedSet}
\newcommand{\measurableRectangle}{\measurableIndexSet \times \measurableInducing}

\newcommand\likelihoodSymbol{L}
\newcommand\generalLikelihood{\likelihoodSymbol(\outputDataSet | \genericFunction) }
\newcommand\generalMarginalLikelihood{\likelihoodSymbol( \outputDataSet )}
\newcommand\dataLikelihood{\likelihoodSymbol(\outputDataSet | \dataFunction )}

\newcommand{\polishSpace}{U}
\newcommand{\polishSpaceB}{V}
\newcommand{\productPolishSpace}{\polishSpace \times \polishSpaceB}

\newcommand{\sigmaAlgebra}{\mathcal{F}}
\newcommand{\sampleSpaceIndexSet}{\genericSet_{\indexSet}}
\newcommand{\sampleSpaceInducing}{\genericSet_{\augmentedSet}}
\newcommand{\sampleSpaceProduct}{\sampleSpaceIndexSet \times \sampleSpaceInducing }

\newcommand{\cartesianProduct}{\times}

\newcommand{\sigmaAlgebraIndexSet}{\sigmaAlgebra_{\indexSet}}
\newcommand{\sigmaAlgebraInducing}{\sigmaAlgebra_{\augmentedSet}}

\newcommand{\indexSetConditionedOnInducingNoSet}[1]{ \priorMeasure_{\indexSet | \augmentedSet} ( #1 ) }
\newcommand{\indexSetConditionedOnInducingDotSet}{\indexSetConditionedOnInducingNoSet{\LargerCdot}}

\newcommand{\inducingSetConditionedOnIndexSetNoSet}[1]{ \priorMeasure_{\augmentedSet | \indexSet} ( #1 ) }
\newcommand{\inducingSetConditionedOnIndexSetDotSet}{\inducingSetConditionedOnIndexSetNoSet{\LargerCdot}}

\newcommand{\approxInducingSetConditionedOnIndexSetNoSet}[1]{ \approximatingMeasure_{\augmentedSet | \indexSet} ( #1 ) }

\newcommand{\isDominatedBy}{ << }

\newcommand{\sigmaAlgebraProduct}{\sigmaAlgebraIndexSet \times \sigmaAlgebraInducing }
\newcommand{\modelMeasure}{P}
\newcommand{\modelMeasureIndexMarginal}{\modelMeasure_\indexSet}
\newcommand{\modelMeasureInducingMarginal}{\modelMeasure_\augmentedSet}

\newcommand{\inverseTransformationFunction}{\transformationFunction^{-1}}

\newcommand{\approximatingMeasureIndexMarginal}{\approximatingMeasure_\indexSet}
\newcommand{\approximatingMeasureInducingMarginal}{\approximatingMeasure_\augmentedSet}

\newcommand{\realLine}{\mathbb{R}}

\newcommand{\radonDerivativeInducing}{ \frac{ \dee \approximatingMeasureInducingMarginal }{ \dee \modelMeasureInducingMarginal } }
\newcommand{\radonDerivativeConditional}{ \frac{ \dee \modelMeasure( \measurableIndexSet \cartesianProduct \LargerCdot ) }{\dee \modelMeasureInducingMarginal( \LargerCdot  ) } }

\newcommand{\interDomainVariable}{i}
\newcommand{\interDomainVariableFunction}{\genericFunction_{\interDomainVariable}}
\newcommand{\smootherFunction}{g}
\newcommand{\interDomainSmoother}{\smootherFunction_\interDomainVariable}
\newcommand{\indexSetMeasure}{\lambda}

\newcommand{\equationFullStop}{\,.}

\newcommand{\projection}{\pi}

\title{  \hsize\textwidth
  \linewidth\hsize \toptitlebar {\centering
  {\Large\bf On Sparse Variational Methods and the Kullback\hyp{}Leibler Divergence between Stochastic Processes }}
 \bottomtitlebar \vskip 0.2in plus 1fil minus 0.1in}

\author[1]{\hspace{3 pt}\fontsize{10.6}{13}{\bf{Alexander G. de G. Matthews}}}
\author[2]{\hspace{-0.5 pt}\fontsize{10.6}{13}{\bf{James Hensman}}}
\author[1]{\hspace{-0.5 pt}\fontsize{10.6}{13}{\bf{Richard E. Turner}}}
\author[1]{\hspace{-1 pt}\fontsize{10.6}{13}{\bf{Zoubin Ghahramani}}}
\affil[1]{University of Cambridge} 
\affil[2]{Lancaster University}

\date{}

\begin{document}

\twocolumn[

\aistatstitle{On Sparse Variational Methods and the Kullback\hyp{}Leibler Divergence between Stochastic Processes}

\aistatsauthor{ Alexander G. de G. Matthews \and James Hensman \and Richard E. Turner \and Zoubin Ghahramani }

\vskip -0.2in 
\maketitle

]




\begin{abstract}
The variational framework for learning inducing variables \citep{Titsias2009} has had a large impact on the Gaussian process literature. The framework may be interpreted as minimizing a rigorously defined Kullback-Leibler divergence between the approximating and posterior processes. To our knowledge this connection has thus far gone unremarked in the literature. In this paper we give a substantial generalization of the literature on this topic. We give a new proof of the result for infinite index sets which allows inducing points that are not data points and likelihoods that depend on all function values. We then discuss augmented index sets and show that, contrary to previous works, marginal consistency of augmentation is not enough to guarantee consistency of variational inference with the original model. We then characterize an extra condition where such a guarantee is obtainable. Finally we show how our framework sheds light on interdomain sparse approximations and sparse approximations for Cox processes.
\end{abstract}

\section{Introduction}

The variational approach to inducing point selection of Titsias \citeyearpar{Titsias2009} has been highly influential in the active research area of scalable Gaussian process approximations.
The chief advantage of this particular framework is that the inducing points positions are variational parameters rather than model parameters and as such are protected from overfitting. In this paper we argue that whilst this is true, it may not be for exactly the reasons previously thought.
The original framework is applied to conjugate likelihoods and has been extended to non-conjugate likelihoods \citep{Chai2012,Hensman2015}. 
An important advance in the use of variational methods was their combination with stochastic gradient descent \citep{Hoffman2013} and the variational inducing point framework has been combined with such methods in the conjugate \citep{Hensman2013} and non-conjugate cases \citep{Hensman2015}. The approach has also been successfully used to perform scalable inference in more complex models such as the Gaussian process latent variable model \citep{Titsias2010,Damianou2014} and the related Deep Gaussian process \citep{Damianou2012,Hensman2014}.

To be more concrete let us set up some notation. Consider a function $\genericFunction$ mapping an index set $\indexSet$ to the set of real numbers  $\genericFunction : \indexSet \mapsto \realNumbers$. Entirely equivalently we may write $ \genericFunction \in \realNumbersOverIndexSet $ or use sequence notation $ \left( \genericFunction( \indexSetMember ) \right)_{\indexSetMember \in \indexSet} $. We also define set indexing of the function. If $\subsetIndexSet \subseteq \indexSet$ is some subset of the index set, then $\genericFunction_{\subsetIndexSet} := \left( \genericFunction(\indexSetMember) \right)_{\indexSetMember \in \subsetIndexSet}$ and we may straightforwardly extend this definition to single elements of the index set $ \genericFunction_{\indexSetMember} := \genericFunction_{\lbrace \indexSetMember \rbrace } $. 
We can put this notation to immediate use by defining a subset $\dataSet \subseteq \indexSet$ of the index set, of size $\nDataPoints$, that corresponds to those input points for which we have observed data. The corresponding function values will then be denoted $\dataFunction$. For simplicity, we will initially assume that we have one, possibly noisy, possibly non-conjugate observation $\outputDataPoint$ per input data point which will together form a set $\outputDataSet$.

Gaussian processes allow us to define a prior over functions $\genericFunction$. After we observe the data we will have some posterior which we wish to approximate with a sparse distribution. At the heart of the variational inducing point approximation is the idea of `augmentation' that appears in the original paper and many subsequent ones.
We choose to monitor a set $ \inducingPointIdentities \subseteq \indexSet $ of size $\nInducingPoints$. These points may have some overlap with the input data points $\dataSet$ but to give a computational speed up $\nInducingPoints$ will need to be less than the number of data points $\nDataPoints$. The Kullback-Leibler divergence given as an optimization criterion in Titsias' original paper is
\begin{align}\label{eq:titsiasKL}
& \KL[ q( \dataFunctionNoInd, \indFunction ) || p( \dataFunctionNoInd, \indFunction | \outputDataSet )  ] \notag \\
= \int &  q( \dataFunctionNoInd, \genericFunction_{\inducingPointIdentities} ) \log  \left\lbrace \frac{ q( \dataFunctionNoInd, \indFunction ) }{ p( \dataFunctionNoInd, \indFunction | \outputDataSet )  } \right\rbrace  \dee \dataFunctionNoInd \dee \indFunction \equationFullStop
\end{align}

The variational distribution at those data points which are not also inducing points is taken to have the form:

\begin{equation}
 q( \dataFunctionNoInd, \indFunction ) := p( \dataFunctionNoInd | \indFunction ) q( \indFunction  )
\end{equation}

\noindent where $p( \dataFunctionNoInd | \indFunction )$ is the prior conditional and $q(\indFunction) $ is a variational distribution on the inducing points only. Under this factorization, for a conjugate likelihood, the optimal $q(\indFunction)$ has an analytic Gaussian solution \citep{Titsias2009}. The non-conjugate case was then studied in subsequent work \citep{Chai2012,Hensman2015}. In both cases the sparse approximation requires only $\bigO( \nDataPoints \nInducingPoints^2) $ rather than the $\bigO( \nDataPoints^3 ) $ required by exact methods in the conjugate case, or many commonly used non-conjugate approximations that don't assume sparsity.

The augmentation is justified by arguing that the model remains marginally the same when the inducing points are added. It is therefore suggested that variational inference in the augmented model, including for the parameters of said augmentation, is equivalent to variational inference in the original model, i.e that the inducing point positions can be considered to be variational parameters and are consequently protected from overfitting. For example see Titsias' original conference paper \citep{Titsias2009}, section 3 or the longer technical report version \citep{Titsias2009b}, section 3.1. In the common case in the literature where the argument proceeds by applying Jensen's inequality to the marginal likelihood as, for example, in Hensman et al \citeyearpar{Hensman2015} equations (6) and (17), the slack of the bound on the marginal likelihood is precisely the $\KL$-divergence \eqref{eq:titsiasKL}. Therefore maximizing such a bound is exactly equivalent to minimizing this objective and the considerations that follow all apply.

In fact in this paper, whilst we applaud the excellent prior work, we will show that variational inference in an augmented model is not equivalent to variational inference in the original model. Without this justification, the $\KL$-divergence in equation \eqref{eq:titsiasKL} could seem to be a strange optimization target. The $\KL$-divergence has the inducing variables on both sides, so it might seem that in optimizing the inducing point positions we are trying to hit a `moving target'. It is desirable to rigorously formulate a `one sided' $\KL$-divergence that leads to Titsias' formulation. Such a derivation could be viewed as putting these elegant and popular methods on a firmer foundation. Such a derivation is the topic of this article. As we shall show this cements the framework for sparse interdomain inducing approximations and sparse variational inference in Cox processes. We wish to re-emphasize our respect for the previous work and for the avoidance of suspense we will find that much of the existing work carries over \emph{mutatis mutandis}. Nevertheless we feel that most readers at the end of the paper will agree that a precise treatment of the topic should be of benefit going forward.

In terms of prior work for the theoretical aspect, the major other references are the early work of Seeger \citeyearpar{Seeger2003b,Seeger2003}. In particular Seeger identifies the $\KL$-divergence \emph{between processes} (more commonly referred to as a relative entropy in those texts) as a measure of similarity and applies it to PAC-Bayes and to subset of data sparse methods. Crucially, Seeger outlines the rigorous formulation of such a $\KL$-divergence which is a large technical obstacle. Here we give a shorter, more general, and intuitive proof of the key theorem.
We extend the stochastic process formulation to inducing points which are not necessarily selected from the data and show that this is equivalent to Titsias' formulation. In so far as we are aware this relationship has not previously been noted in the literature. The idea of using the $\KL$-divergence between processes is also mentioned in the early work of Csato and Opper \citeyearpar{Csato2002,Csato2002b} but the transition from finite dimensional multivariate Gaussians to infinite dimensional Gaussian processes is not covered at the level of detail discussed here. An optimization target that in intent seems to be similar to a $\KL$-divergence between stochastic process is briefly mentioned in the work of Alvarez \citeyearpar{Alvarez2011}. The notation used suggests that the integration is with respect to an `infinite dimensional Lebesgue measure', which as we shall see is an argument that arrives at the right answer via a mathematically flawed route. Chai \citeyearpar{Chai2012} seems to have been at least partly aware of Seeger's $\KL$-divergence theorems \citep{Seeger2003} but instead uses them to bound the finite joint predictive probability of a non sparse process.

This article proceeds by first discussing the finite dimensional version of the full argument. This requires considerably less mathematical machinery and much of the intuition can be gained from this case. We then proceed to give the full measure theoretic formulation, giving a new proof that allows inducing points that are not data points and for the likelihood to depend on infinitely many function values. Next we discuss augmentation of the original index set, using the crucial chain rule for $\KL$-divergences. This gives us a framework to discuss marginal consistency and how variational inference in augmented models is not necessarily equivalent to variational inference in the original model. We then show that under very general conditions augmentation which is deterministic conditioned on the whole latent function does have the desired property. We apply our results to sparse variational interdomain approximations and to posterior inference in Cox processes. Finally we conclude and highlight avenues for further research.

\section{Finite index set case}\label{section:finiteIndexSet}

This section is in fact a less general case of what follows. It is included for the benefit of those familiar with the previous work on variational sparse approximations and as an important special case. Consider the case where $\indexSet$ is finite. We introduce a new set $ \holdOutSet := X \backslash ( \dataSet \cup \inducingPointIdentities ) $, in words: all points that are in the index set that aren't inducing points or data points. These points might be of practical interest for instance when making predictions on hold out data.

We extend the variational distribution to include these points:
\begin{equation}
 q( \holdOutFunction , \dataFunctionNoInd, \indFunction ) := p( \holdOutFunction , \dataFunctionNoInd | \indFunction ) q( \indFunction ).
\end{equation}
We then consider the $\KL$-divergence between this extended variational distribution and the full posterior distribution $ p( \genericFunction | \outputDataSet ) $
\begin{align}\label{eq:finiteKL}\
&\KL[  q( \holdOutFunction , \dataFunctionNoInd, \indFunction ) || p( \genericFunction | \outputDataSet ) ] \notag \\
=& \KL[  q( \holdOutFunction , \dataFunctionNoInd, \indFunction ) || p( \holdOutFunction , \dataFunctionNoInd, \indFunction | \outputDataSet ) ] \notag \\
=& \int q( \holdOutFunction , \dataFunctionNoInd, \indFunction ) \log \frac{q( \holdOutFunction , \dataFunctionNoInd, \indFunction ) }{p( \holdOutFunction , \dataFunctionNoInd, \indFunction | \outputDataSet ) } \dee \holdOutFunction \dee \dataFunctionNoInd \dee \indFunction
\end{align}
Next we expand the term inside the logarithm and cancel one of the terms that appears in both the numerator and the denominator:
\begin{align}\label{eq:fractions}
&\frac{q( \holdOutFunction , \dataFunctionNoInd, \indFunction ) }{p( \holdOutFunction , \dataFunctionNoInd, \indFunction | \outputDataSet ) } \notag \\
=&\frac{ p(\holdOutFunction | \dataFunctionNoInd, \indFunction ) p( \dataFunctionNoInd | \indFunction ) q(\indFunction) p(\outputDataSet) }{ p(\holdOutFunction | \dataFunctionNoInd, \indFunction ) p( \dataFunctionNoInd | \indFunction ) p( \indFunction ) p(\outputDataSet | \dataFunction) } \notag \\
=&\frac{ p( \dataFunctionNoInd | \indFunction ) q(\indFunction) p(\outputDataSet) }{ p( \dataFunctionNoInd | \indFunction ) p( \indFunction ) p(\outputDataSet | \dataFunction) } \notag \\
=&\frac{q( \dataFunctionNoInd, \indFunction )}{p(\dataFunctionNoInd, \indFunction | \outputDataSet )}
\end{align}
Substituting back into the full integral and exploiting the marginalization property of the conditional density we obtain:
\begin{align}
&\int p( \holdOutFunction , \dataFunctionNoInd | \indFunction ) q(\indFunction) \log \frac{q( \dataFunctionNoInd, \indFunction )}{p(\dataFunctionNoInd, \indFunction | \outputDataSet )} \dee \holdOutFunction \dee \dataFunctionNoInd \dee \indFunction \notag \\
&=\int p( \dataFunctionNoInd | \indFunction ) q(\indFunction) \log \frac{q( \dataFunctionNoInd, \indFunction )}{p(\dataFunctionNoInd, \indFunction | \outputDataSet )} \dee \dataFunctionNoInd \dee \indFunction
\end{align}

The last line is exactly the $\KL$-divergence used by Titsias \citeyearpar{Titsias2009} that we already described in equation \eqref{eq:titsiasKL}. We thus see that for finite index sets considering the $\KL$-divergence between the two distributions is equivalent to Titsias' $\KL$-divergence. We might choose to optimize our choice of the $\nInducingPoints$ by selecting them from the $|\indexSet|$ possible values in the index set and comparing the $\KL$-divergence between distributions given in equation \eqref{eq:finiteKL}. The equivalence with equation \eqref{eq:titsiasKL} that we have just derived shows us that in this case the appearance of the inducing values on both sides of the equation is just a question of `accounting'. That is to say, whilst we are in fact optimizing the $\KL$-divergence between the full distributions, we only need to keep track of the distribution over function values $\indFunction$ and $\dataFunctionNoInd$. All the other function values $\holdOutFunction$ marginalize. For different choices of inducing points we will need to keep track of different function values and be able to safely ignore different values $\holdOutFunction$. 

\section{Infinite index set case}\label{section:infiniteIndexSet}

\subsection{There is no useful infinite dimensional Lebesgue measure}{\label{section:noInfiniteLebesgueMeasure} 

One might hope to cope with not only finite index sets but also infinite index sets in the way discussed in section \ref{section:finiteIndexSet}. Unfortunately when $\indexSet$ and hence $\holdOutFunction$ are infinite sets we cannot integrate with respect to a `infinite dimensional vector'. That is to say the notation  $\int (\cdot) \dee \holdOutFunction $ can no longer be correctly used.

For a discussion of this see, for example, Hunt et al \citeyearpar{Hunt1992}. The crux of the issue is that to give sensible answers such a measure would need to be translation invariant and locally finite. Unfortunately the only measure that obeys these two properties is the zero measure which assigns zero to every input set. 

Thus we see that it will be necessary to rethink our approach to a $\KL$-divergence between stochastic processes. It will turn out that a reasonable definition will require the full apparatus of measure theory. Readers looking for some background on these issues may wish to consult a larger text \citep{Billingsley1995,Capinski2004}.

\subsection{The $\KL$-divergence between processes}

In this section we review the rigorous definition of the $\KL$-divergence between stochastic processes \citep{Gray2011}. 

Suppose we have two measures $\genericMeasure$ and $\genericMeasureB$ for $(\genericSet,\genericSigmaAlgebra)$ and that $\genericMeasure$ is absolutely continuous with respect to $\genericMeasureB$. Then there exists a Radon-Nikodyn derivative $\genericRadonNikodymDerivative$ and the correct definition for $\KL$-divergence between these measures is:
\begin{equation}
\KL[ \genericMeasure || \genericMeasureB] = \int_{\genericSet} \log \left\lbrace \genericRadonNikodymDerivative \right\rbrace \dee \genericMeasure \equationFullStop
\end{equation}
In the case where $\genericMeasure$ is not absolutely continuous with respect to $\genericMeasureB$ we let $\KL[ \genericMeasure || \genericMeasureB] =\infty $. In the case where the sample space is $\realLine^K$ for some finite $K$ and both measures are dominated by Lebesgue measure $m$ this reduces to the more familiar definition:
\begin{equation}
\KL[ \genericMeasure || \genericMeasureB] = \int_{\genericSet} u \log \left\lbrace \frac{u}{v} \right\rbrace \dee m
\end{equation}
\noindent where $u$ and $v$ are the respective densities with respect to Lebesgue measure.  The first definition is more general and allows us to deal with the problem of there being no sensible infinite dimensional Lebesgue measure by instead integrating with respect to the measure $\genericMeasure$.

\subsection{A general derivation of the sparse inducing point framework}\label{section:generalInducingPointFramework}

In this section we give a general derivation of the sparse inducing point framework. The derivation is more general than that of Seeger \citeyearpar{Seeger2003b,Seeger2003} since it does not require that the inducing points are selected from the data points. Nor does it assume that the relevant finite dimensional marginal distributions have density with respect to Lebesgue measure. Finally since the dependence on the elegant properties of Radon-Nikodym derivatives has been made more explicit we believe it is clearer \emph{why} the derivation works and how one would generalize it.

We are now interested in three types of probability measure on sets of functions $\genericFunction : \indexSet \mapsto \mathbb{R} $. The first is the prior measure $\priorMeasure$ which will be assumed to be a Gaussian process. The second is the approximating measure $\approximatingMeasure$ which will be assumed to be a sparse Gaussian process and the third is the posterior process $\posteriorMeasure$ which may be Gaussian or non-Gaussian depending on whether we have a conjugate likelihood. We start with a measure theoretic definition of Bayes' theorem for a dominated model \citep{Schervish1995}. It specifies the Radon-Nikodym derivative of the posterior with respect to the prior.
\begin{equation}\label{eq:generalBayesTheorem}
\bayesRadonNikodymDerivative(\genericFunction) = \frac{\generalLikelihood}{\generalMarginalLikelihood}
\end{equation}
\noindent with $\generalLikelihood$ being the likelihood and $\generalMarginalLikelihood = \int_{\realNumbersOverIndexSet} \generalLikelihood \dee \priorMeasure( \genericFunction )$ the marginal likelihood. As we have assumed in previous sections we will initially restrict the likelihood to only depend on the finite data subset of the index set. We denote by $\projection_\projectionSet : \realNumbersOverIndexSet \mapsto \realLine^\projectionSet $ a projection function, which takes the whole function as an argument and returns the function at some set of points $\projectionSet$. In this case we have:
\begin{equation}\label{eq:dataBayesTheorem}
\bayesRadonNikodymDerivative(\genericFunction) = \bayesRadonNikodymDerivativeMarginal( \projection_\dataSet(\genericFunction ) ) = \frac{\likelihoodSymbol( \outputDataSet | \projection_\dataSet(\genericFunction ) )}{\generalMarginalLikelihood}
\end{equation}
\noindent and similarly the marginal likelihood only depends on the function values on the data set $ \generalMarginalLikelihood = \int_{\realNumbersOverDataSet} \dataLikelihood \dee \priorMeasure_\dataSet( \dataFunction ) $. In fact, we will relax the assumption that the data set is finite in section \ref{section:coxProcesses} and the ability to do so is one of the benefits of this framework. Next we specify $\approximatingMeasure$ by assuming it has density with respect to the posterior and thus the prior and that the density with respect to the prior depends on some set of points $\inducingPointIdentities$: 
\begin{equation}\label{eq:approximation}
\approximatingRadonNikodyDerivative(\genericFunction) = \approximatingRadonNikodyDerivativeMarginal( \projection_\inducingPointIdentities(\genericFunction) ) \equationFullStop
\end{equation}
\noindent Under this assumption $\approximatingMeasure$ is fully specified if we know $\priorMeasure$ and $\approximatingRadonNikodyDerivativeMarginal$. To gain some intuition for this assumption we can compare equations \eqref{eq:approximation} and \eqref{eq:dataBayesTheorem}. We see that in the approximating distribution the set $\inducingPointIdentities$ is playing a similar one to that played for $\dataSet$ in the true posterior distribution. We now bring these assumptions together. Let us apply the chain rule for Radon-Nikodym derivatives and a standard property of logarithms:
\begin{align}\label{eq:chainRuleRadonNikodym}
&\KL[ \approximatingMeasure || \posteriorMeasure ] \notag \\
= &\int_{\realNumbersOverIndexSet} \log \left\lbrace \approximatingRadonNikodyDerivative( \genericFunction ) \right \rbrace \dee \approximatingMeasure( \genericFunction ) - \int_{\realNumbersOverIndexSet} \log \left\lbrace \bayesRadonNikodymDerivative( \genericFunction ) \right \rbrace \dee \approximatingMeasure( \genericFunction ) \equationFullStop
\end{align}
Taking the first term alone we exploit the sparsity assumption for the approximating distribution:
\begin{align}
&\int_{\realNumbersOverIndexSet} \log \left\lbrace \approximatingRadonNikodyDerivative( \genericFunction ) \right \rbrace \dee \approximatingMeasure( \genericFunction ) \notag \\
=&\int_{\realNumbersOverInducingSet} \log \left\lbrace \approximatingRadonNikodyDerivativeMarginal( \indFunction ) \right \rbrace \dee \approximatingMeasure_\inducingPointIdentities( \indFunction ) \equationFullStop
\end{align}
Taking the second term in the last line of equation \eqref{eq:chainRuleRadonNikodym} and exploiting the measure theoretic Bayes' theorem we obtain:
\begin{align}
&\int_{\realNumbersOverIndexSet} \log \left\lbrace \bayesRadonNikodymDerivative( \genericFunction ) \right \rbrace \dee \approximatingMeasure( \genericFunction ) \notag \\
= &\int_{\realNumbersOverDataSet} \log \left\lbrace \bayesRadonNikodymDerivativeMarginal( \dataFunction ) \right \rbrace \dee \approximatingMeasure_\dataSet( \dataFunction ) \notag \\
=& \expectation_{\approximatingMeasure_\dataSet}\left[ \log \dataLikelihood \right] - \log \generalMarginalLikelihood \equationFullStop
\end{align}
Finally noting the appearance of a marginal $\KL$-divergence we obtain our result:
\begin{equation}\label{eq:finalResult}
\KL[ \approximatingMeasure || \posteriorMeasure ] = \KL[ \approximatingMeasure_\inducingPointIdentities || \priorMeasure_\inducingPointIdentities ] - \expectation_{\approximatingMeasure_\dataSet}\left[ \log \dataLikelihood \right] + \log \generalMarginalLikelihood \equationFullStop
\end{equation}
\noindent As is common with variational approximations, in most cases of interest the marginal likelihood will be intractable. However since it is an additive constant, independent of $\approximatingMeasure$, it can be safely ignored. The final equation shows that we need to be able to compute the $\KL$-divergence between the inducing point marginals of the approximating distribution and the prior for all $ \inducingPointIdentities \subset \indexSet $ and the expectation under the data marginal distribution of $\approximatingMeasure$ of the log likelihood. In the case where the likelihood factorizes across data terms this will give a sum of one dimensional expectations. Note the similarity of equation \eqref{eq:finalResult} with \cite{Hensman2015} equation (17) where a less general expression is motivated from a `model augmentation' view. \!Notice that at no point in our derivation did we try to envoke the pathological `infinite dimensional Lebesgue measure' which is important for the reasons discussed in section \ref{section:noInfiniteLebesgueMeasure}. The ease of derivation suggests that Radon-Nikodym derivatives and measure theory provide the most natural and general way to think about such approximations.

\section{Augmented index sets}

We now consider the case where we supplement the original (finite or infinite) index set $\indexSet$ with a finite set of elements $\augmentedSet$, intending to use them as inducing points. The precise nature of the augmented prior model will be parameterized by some parameters $\augmentationParameters$ which we will hope to tune to give a good approximation. It will be seen that the this is very much in the spirit of the original augmentation argument given by Titsias \citeyearpar{Titsias2009} and the `variational compression' framework of Hensman and Lawrence \citeyearpar{Hensman2014}. \noindent This setup also covers the case of variational `interdomain' Gaussian processes which were mooted but not implemented in Figueiras-Vidal and Lazaro-Gredilla \citeyearpar{Figueiras2009} and implemented under the basis of the marginal consistency argument in Alvarez et al \citeyearpar{Alvarez2011b}. We intend to discuss the marginal consistency argument in some detail and we shall deal with the thorny issues surrounding the rigorous treatment of the various infinities involved.

Marginal consistency is easily ensured by specifying the distribution of the augmented function value points $\augmentedSetFunction$ conditioned on the values of the function on the original set $\indexSetFunction$. We denote the corresponding measure as $ \priorMeasure_{\augmentedSet | \indexSet}( \LargerCdot \hspace{2 pt};\hspace{2 pt} \augmentationParameters ) $\footnote{Note that for brevity our notation for conditional measures won't include the explicit function dependence. For example, in this case we omit the explicit dependence on $\indexSetFunction$. }. Let $\genericSet_\indexSet = \realNumbersOverIndexSet$ and $\genericSet_\augmentedSet = \realNumbersOverAugmentedSet$ be the sample spaces associated with the original index set and the augmenting variables respectively. Let $\sigmaAlgebraIndexSet$ and $\sigmaAlgebraInducing$ be their $\sigma$-algebras. Marginal consistency states that we will be interested in probability measures that have the following behaviour on the measurable rectangles $\measurableRectangle \in  \sigmaAlgebraProduct$:
\begin{equation}
\priorMeasure_{\overallSet}( \measurableRectangle  ; \augmentationParameters) = \int_{A_{\indexSet}} \priorMeasure_{\augmentedSet | \indexSet}( A_{\augmentedSet} ; \augmentationParameters ) \dee \priorMeasure_{\indexSet}( \indexSetFunction ).
\end{equation}
We have included the augmentation parameters $\augmentationParameters$ explicitly up until now, but for brevity we will omit them in what follows. We will make this marginal consistency assumption in all that follows. Let us call the overall set $\overallSet$ the `union set'. In a similar vein to the previous section we assume that the approximating measure $\approximatingMeasureOverallSet$ has density with respect to the augmented prior model $\priorMeasureOverallSet$ and that the Radon-Nikodym derivative is only a function of the augmented function points:
\begin{equation}
\approximatingRadonNikodyDerivativeOverallSet( \overallSetFunction ) = \approximatingRadonNikodyDerivativeMarginalAugmented( \projection_{\augmentedSet} ( \overallSetFunction ) ) \equationFullStop
\end{equation}
\noindent Acting as if the augmented set were the original index set we would obtain by a similar argument:
\begin{align}
\KL[ \approximatingMeasureOverallSet || \posteriorMeasureOverallSet ] = \hspace{7 pt} &\KL[ \approximatingMeasure_\augmentedSet || \priorMeasure_\augmentedSet ] \notag \\
- &\expectation_{\approximatingMeasure_\dataSet}\left[ \log \dataLikelihood \right] + \log \generalMarginalLikelihood \equationFullStop
\end{align}
\noindent Sharp eyed readers, however, will have noted that since $\posteriorMeasureOverallSet$ depends on the augmentation parameters $\augmentationParameters$ we are back in a situation where we can tune the approximation on the left hand side and the optimization target on the right. As we will see in the next section we are not necessarily rescued by the marginal consistency argument. It is not the case in general that $\KL[ \approximatingMeasure_\indexSet || \posteriorMeasure_\indexSet ] $ equals $\KL[ \approximatingMeasureOverallSet || \posteriorMeasureOverallSet ]$. In fact the relationship is governed by the chain rule for $\KL$-divergences as we shall now see.

\subsection{The chain rule for $\KL$-divergences}

For what follows we will require the chain rule for $\KL$-divergences \citep{Gray2011}. Let $\polishSpace$ and $\polishSpaceB$ be two Polish spaces endowed with their standard Borel $\sigma$-algebras and let $\productPolishSpace$ be the Cartesian product of these spaces endowed with the corresponding product $\sigma$-algebra. Consider two probability measures $\genericMeasure_{\productPolishSpace}, \genericMeasureB_{\productPolishSpace}$ on this product space and let $\genericMeasure_{\polishSpace |\polishSpaceB}, \genericMeasureB_{\polishSpace |\polishSpaceB} $ be the corresponding regular conditional measures. Assume that $\genericMeasure_{\productPolishSpace}$ is dominated by $\genericMeasureB_{\productPolishSpace}$. The chain rule for $\KL$-divergences says that:
\begin{equation}
\KL[\genericMeasure_{\productPolishSpace}||\genericMeasureB_{\productPolishSpace} ]\!=\expectation_{\genericMeasure_{\polishSpaceB\!}}\!\left\lbrace \KL[\hspace{1 pt}\genericMeasure_{\polishSpace |\polishSpaceB} || \genericMeasureB_{\polishSpace |\polishSpaceB}  ]\hspace{0.5 pt} \right\rbrace +\KL[ \genericMeasure_{\polishSpaceB} || \genericMeasureB_{\polishSpaceB} ].
\end{equation}
The first term on the right hand side is referred to as the `conditional $\KL$-divergence' or `conditional relative entropy'.

\subsection{The marginally consistent augmentation argument is not correct in general.}

Applying the chain rule for $\KL$-divergences to the divergence on the union set we obtain:
\begin{align}\label{eq:chainRuleForAugmentedKL}
&\hspace{6 pt}\KL[ \approximatingMeasureOverallSet || \posteriorMeasureOverallSet ] \notag \\
= \expectation_{\hspace{1 pt}\approximatingMeasure_{\indexSet}} &\left\lbrace \KL[\hspace{2 pt}\approximatingMeasure_{\augmentedSet |\indexSet} || \posteriorMeasure_{\augmentedSet |\indexSet}  ]\hspace{2 pt} \right\rbrace + \KL[ \approximatingMeasure_{\indexSet} || \posteriorMeasure_{\indexSet} ] \notag \\
= \expectation_{\hspace{1 pt}\approximatingMeasure_{\indexSet}} &\left\lbrace \KL[\hspace{2 pt}\approximatingMeasure_{\augmentedSet |\indexSet} || \priorMeasure_{\augmentedSet |\indexSet}  ]\hspace{2 pt} \right\rbrace + \KL[ \approximatingMeasure_{\indexSet} || \posteriorMeasure_{\indexSet} ] \equationFullStop
\end{align}
\noindent The final line follows from the fact that in the assumed model augmentation scheme the additional variables $\augmentedSetFunction$ are conditionally independent of the data given $\indexSetFunction$. This relation makes precise our claim that marginal consistency is not enough to guarantee that $\KL[ \approximatingMeasure_\indexSet || \posteriorMeasure_\indexSet ]$ equals $\KL[ \approximatingMeasureOverallSet || \posteriorMeasureOverallSet ]$. In fact this will only be true if $\approximatingMeasure_{\augmentedSet |\indexSet} = \priorMeasure_{\augmentedSet |\indexSet} $, $\approximatingMeasure_{\indexSet}$-almost surely. In the case where this is not true variational inference in the family of augmented models is not equivalent to variational inference in the original model and we will be optimizing a `two-sided' objective function. We will consider an important condition which ensures the desired equality does hold in the next section.

Before we move on, however, it is also instructive to consider a transformation of the original unaugmented problem into the augmented problem. Take the transformed augmentation set and index set $( \tilde{\augmentedSet},\tilde{\indexSet} ) $ to be defined in terms of the old sets as $(\indexSet \backslash \dataSet, \dataSet)$. The chain rule then tells us that the $\KL$-divergence on the data set is not in general equal to the $\KL$-divergence on the index set although this is true if $\inducingPointIdentities \subset \dataSet$.
\subsection{Deterministic augmentation}{\label{section:deterministicAugmentation}}
Here we discuss an important case where the augmented $\KL$-divergence and the unaugmented $\KL$-divergence are indeed equal, namely where the additional variables $\augmentedSetFunction$ are a deterministic function $\transformationFunction$ of the function values on the original index set $\indexSetFunction$. A few conceptual points may be useful before we go into the detail. First the constraint only says that the values are deterministic conditioned on the function over the whole index set and the index set itself may be infinite. Usually in practice either through noise, finite observations or both, we can't know the latent function exactly and hence in our model we won't know the inducing variables exactly. Second, whilst this assumption may initially seem contrived, in fact it covers two very important cases: the original framework where some inducing points are selected from the index set $\indexSet$ then `copied' over to $\augmentedSet$ and as we shall see later the interdomain inducing point framework.

Having a deterministic function mapping is equivalent to having a delta function conditional distribution centred on the function value. Thus the conditional $\KL$-divergence term in equation \eqref{eq:chainRuleForAugmentedKL} i.e the expectation of the conditional on the right hand side, will be zero if the approximating measure $\approximatingMeasureOverallSet$ has the same delta function conditional. The next theorem shows that if we follow the usual prescription for defining $\approximatingMeasureOverallSet$ this will indeed be the case.

\subsubsection{The governing theorem on deterministic augmentation}

Let $(\sampleSpaceIndexSet,\sigmaAlgebraIndexSet)$ and $(\sampleSpaceInducing,\sigmaAlgebraInducing)$ be two Polish spaces and let $(\sampleSpaceProduct,\sigmaAlgebraProduct)$ be their product space endowed with product $\sigma$-algebra. Let $\transformationFunction : \sampleSpaceIndexSet \mapsto \sampleSpaceInducing $ be a $\sigmaAlgebraIndexSet \slash \sigmaAlgebraInducing$ measurable function. We are interested in a measure $\modelMeasure : \sigmaAlgebraProduct \mapsto \realLine $ which has the following property on the measurable rectangles $\measurableRectangle$
\begin{equation}\label{eq:modelMeasureProperty}
\modelMeasure( \measurableRectangle ) = \modelMeasureIndexMarginal( \measurableIndexSet \cap \inverseTransformationFunction( \measurableInducing ) )
\end{equation}
\noindent where $ \modelMeasureIndexMarginal := \modelMeasure( \measurableIndexSet \cartesianProduct \sampleSpaceInducing ) $ is the marginal distribution for $\indexSet$. This assumption in turn implies that the marginal distribution for $\augmentedSet$ has the form
\begin{equation}
\modelMeasureInducingMarginal( \measurableInducing ) = \modelMeasureIndexMarginal( \inverseTransformationFunction( \measurableInducing ) )
\end{equation}
\noindent which is the \emph{push forward measure} of $\modelMeasureIndexMarginal$ under the function $\transformationFunction$. It is clear that the regular conditional distribution $\inducingSetConditionedOnIndexSetDotSet$ has a point measure property:
\begin{equation}
\inducingSetConditionedOnIndexSetNoSet{\measurableInducing} = \delta_{\transformationFunction( \indexSetFunction )}(\measurableInducing) \equationFullStop
\end{equation}
Let $\indexSetConditionedOnInducingDotSet$ be the regular conditional distribution of $\indexSetFunction$ conditioned on $\augmentedSetFunction$. 
Next we define a second measure $\approximatingMeasure : \sigmaAlgebraProduct \mapsto \realLine  $ which has the following property on measurable rectangles
\begin{equation}\label{eq:modelPointMeasureProperty}
\approximatingMeasure( \measurableRectangle ) = \int_{\measurableInducing} \indexSetConditionedOnInducingNoSet{\measurableIndexSet} \dee \approximatingMeasureInducingMarginal( \augmentedSetFunction ) \equationFullStop
\end{equation}
\noindent Finally we assume that $\approximatingMeasureInducingMarginal \isDominatedBy \modelMeasureInducingMarginal $. The theorem states that under the assumptions of the previous section the marginal distributions of $\approximatingMeasure$ have the following property:
\begin{equation}\label{eq:mainTheorem}
\approximatingMeasureInducingMarginal( \measurableInducing ) = \approximatingMeasureIndexMarginal( \inverseTransformationFunction( \measurableInducing ) ) \equationFullStop
\end{equation}
\noindent That is to say the marginal distribution of $\approximatingMeasure$ for $\inducingPointIdentities$ is the push forward measure of  $\approximatingMeasureIndexMarginal$ under the function $\transformationFunction$. Consequently the approximating distribution for $\augmentedSetFunction$ conditioned on $\indexSetFunction$ also has the point measure property
\begin{equation}
\approxInducingSetConditionedOnIndexSetNoSet{\measurableInducing} = \delta_{\transformationFunction( \indexSetFunction )}(\measurableInducing) \equationFullStop
\end{equation}
\noindent We now give a proof. Starting from the right hand side of equation \eqref{eq:mainTheorem}
\begin{align}
\approximatingMeasureIndexMarginal( \inverseTransformationFunction( \measurableInducing ) ) &= \approximatingMeasure( \inverseTransformationFunction( \measurableInducing ) \cartesianProduct \sampleSpaceInducing ) \notag \\
&= \int_{\sampleSpaceInducing} \indexSetConditionedOnInducingNoSet{\inverseTransformationFunction(\measurableInducing)} \dee \approximatingMeasureInducingMarginal( \augmentedSetFunction ) \equationFullStop
\end{align}
\noindent Next since $\approximatingMeasureInducingMarginal \isDominatedBy \modelMeasureInducingMarginal$ we apply the Radon-Nikodym theorem: 
\begin{align}\label{eq:progressSoFar}
\int_{\sampleSpaceInducing} \! \! \indexSetConditionedOnInducingNoSet{\inverseTransformationFunction(\measurableInducing)} \dee \approximatingMeasureInducingMarginal( \augmentedSetFunction )\! =\! \int_{\sampleSpaceInducing} \! \! \indexSetConditionedOnInducingNoSet{\measurableIndexSet}  \radonDerivativeInducing \dee \modelMeasureInducingMarginal( \augmentedSetFunction ) \equationFullStop 
\end{align}
\noindent The existence of conditional distributions is also guaranteed by the Radon-Nikodym theorem. Explicitly we have
\begin{equation}
\indexSetConditionedOnInducingNoSet{\measurableIndexSet} = \radonDerivativeConditional \equationFullStop
\end{equation} 
\noindent Continuing on from equation \eqref{eq:progressSoFar} and applying an elementary theorem of Radon-Nikodym derivatives we have:
\begin{align}
&\int_{\sampleSpaceInducing} \indexSetConditionedOnInducingNoSet{ \inverseTransformationFunction(\measurableInducing) } \radonDerivativeInducing \dee \modelMeasureInducingMarginal( \augmentedSetFunction ) \notag \\
= &\int_{\sampleSpaceInducing} \radonDerivativeInducing \dee \modelMeasure( \inverseTransformationFunction( \measurableInducing) \cartesianProduct \augmentedSetFunction ) \equationFullStop
\end{align}
\noindent Now we apply the property given by equation \eqref{eq:modelMeasureProperty}  
\begin{align}
&\int_{\sampleSpaceInducing} \radonDerivativeInducing \dee \modelMeasure( \inverseTransformationFunction( \measurableInducing) \cartesianProduct \augmentedSetFunction ) \notag \\
= &\int_{\sampleSpaceInducing} \radonDerivativeInducing \dee \modelMeasureIndexMarginal( \inverseTransformationFunction( \measurableInducing) \cap \inverseTransformationFunction( \augmentedSetFunction ) ) \equationFullStop
\end{align}
\noindent Now we apply some algebraic manipulations of the integral:
\begin{align}
&\int_{\sampleSpaceInducing} \radonDerivativeInducing \dee \modelMeasureIndexMarginal( \inverseTransformationFunction( \measurableInducing) \cap \inverseTransformationFunction( \augmentedSetFunction ) ) \notag \\
= &\int_{\sampleSpaceInducing} \radonDerivativeInducing \dee \modelMeasureIndexMarginal( \inverseTransformationFunction( \measurableInducing \cap \augmentedSetFunction ) )  \notag \\
= &\int_{\sampleSpaceInducing} \radonDerivativeInducing \dee \modelMeasureInducingMarginal( \measurableInducing \cap \augmentedSetFunction ) \notag \\
= &\int_{\measurableInducing} \radonDerivativeInducing \dee \modelMeasureInducingMarginal( \augmentedSetFunction ) = \approximatingMeasureInducingMarginal( \measurableInducing )
\end{align}
\noindent as was claimed.

\section{Examples}

\subsection{Variational interdomain approximations}\label{section:interDomain}

\noindent Here we consider the sparse variational interdomain approximation which was suggested but not realized in Figueiras-Vidal and Lazaro-Gredilla \citeyearpar{Figueiras2009} and appeared under the basis of the marginal consistency argument in Alvarez et al \citeyearpar{Alvarez2011b}. An interdomain variable is a random variable, indexed by $\interDomainVariable \in \augmentedSet$ defined in the following way:
\begin{equation}\label{eq:interDomainEquation}
\interDomainVariableFunction(\augmentationParameters) = \int_{X} \interDomainSmoother(\indexSetMember,\augmentationParameters) \indexSetMemberFunction \hspace{2 pt}\dee \indexSetMeasure( \indexSetMember )
\end{equation}
\noindent Here $\indexSetMeasure$ is a measure on $\indexSet$ with some appropriate $\sigma$-algebra, $\lbrace \interDomainSmoother : \interDomainVariable \in \augmentedSet \rbrace$  is a set of $\indexSetMeasure$-integrable functions from $\indexSet$ to $\realLine$. The interdomain variables may be viewed as deterministic conditional on the whole function $\indexSetFunction$ so the theorems of section \ref{section:deterministicAugmentation} come into play. Since the intention here is to put this framework on a firm logical footing, we should also consider the thorny issue of the measurability of this transformation and the associated random variable. The existence of separable measurable versions of stochastic processes including most commonly used Gaussian processes was settled in the work of Doob \citeyearpar{Doob1953}. It also discusses the conditions necessary to apply Fubini's theorem to expectations of the random variable defined by equation \eqref{eq:interDomainEquation}. The application of Fubini's theorem is essential to the utility of such methods in practice \citep{Figueiras2009}.

Thus we may correctly optimize the parameters $\augmentationParameters$ of interdomain inducing points safe in the knowledge that this decision is variationally protected from overfitting and optimizes a well defined $\KL$-divergence objective. The potential for a wide variety of improved sparse approximations in this direction is thus, in our opinion, significant.

\subsection{Approximations to Cox process posteriors}\label{section:coxProcesses}

In this section we relax the assumption that the data set $\dataSet$ is finite, which is necessary to consider Gaussian process based Cox processes. One specific case of this model is considered by Lloyd et al \citeyearpar{lloyd2015} under the marginal consistency motivation. A Gaussian process based Cox process has the following generative scheme:
\begin{align}
\genericFunction &\sim \mathcal{GP}(m,K) \notag \\
h &= \rho( \genericFunction ) \notag \\
\outputDataSet | h &\sim \mathcal{PP}(h) \equationFullStop
\end{align}
\noindent Here $\mathcal{GP}(m,K)$ denotes a Gaussian process with mean $m$ and kernel $K$, $\rho : \realLine \mapsto (0, \infty) $ is an inverse link function, $\mathcal{PP}(h)$ is a Poisson process with intensity $h$ and $\dataSet$ is a set of points in the original index set $\indexSet$. For example in a geographical spatial statistics application we might take $\indexSet$ to be some bounded subset of $\realLine^2 $. The key issue with the Poisson process likelihood is that it depends not just on those points in $\indexSet$ where points where observed but in fact on all points in $\indexSet$. Intuitively the absence of points in an area suggests that the intensity is lower there. Thus $\dataSet = \indexSet$. The likelihood in question is:
\begin{equation}
\dataLikelihood = \left( \prod_{\outputDataPoint \in \outputDataSet} \rho( \outputDataPoint ) \right) \exp \left \lbrace -\int_{\indexSet} \rho( \indexSetMember ) \dee \lebesgueMeasure( \indexSetMember ) \right \rbrace \equationFullStop 
\end{equation}
\noindent where $\lebesgueMeasure$ denotes for instance Lebesgue measure on $\indexSet$. The full $\indexSet$ dependence manifests itself through the integral on the right hand side. We will require that the integral exists almost surely. In Lloyd et al \citeyearpar{lloyd2015} equation (3), the application of Bayes' theorem appears to require a density with respect to infinite dimensional Lebesgue measure. As pointed out in \ref{section:noInfiniteLebesgueMeasure} such a notion is pathological. This however can be fixed because the more general form of Bayes' theorem in equation \eqref{eq:generalBayesTheorem} of this paper still applies. Thus we can apply the results of section \ref{section:generalInducingPointFramework} to obtain:
\begin{align}
\KL[ \approximatingMeasure || \posteriorMeasure ] = &\KL[ \approximatingMeasure_\inducingPointIdentities || \priorMeasure_\inducingPointIdentities ] - \sum_{\outputDataPoint \in \outputDataSet} \expectation_{\approximatingMeasure_{\outputDataPoint}}\left[ \log \rho(\outputDataPoint)\right] \notag \\
&+ \expectation_{\approximatingMeasure_{\indexSet}}\left[\int_{\indexSet} \rho( \indexSetMember ) \dee \lebesgueMeasure( \indexSetMember ) \right] + \log \generalMarginalLikelihood \equationFullStop
\end{align}
\noindent As in section \ref{section:interDomain} we will need to check that the conditions for Fubini's theorem apply \citep{Doob1953} which gives:
\begin{align}
\KL[ \approximatingMeasure || \posteriorMeasure ] = &\KL[ \approximatingMeasure_\inducingPointIdentities || \priorMeasure_\inducingPointIdentities ] - \sum_{\outputDataPoint \in \outputDataSet}\expectation_{\approximatingMeasure_{\outputDataPoint}}\left[ \log \rho(\outputDataPoint)\right] \notag \\
&+ \int_{\indexSet} \expectation_{\approximatingMeasure_\indexSetMember} \left[ \rho(\indexSetMember) \right] \dee \lebesgueMeasure( \indexSetMember ) + \log \generalMarginalLikelihood \equationFullStop
\end{align}
\noindent For the specific case of $\rho$ used in Lloyd et al \citeyearpar{lloyd2015} the working then continues as in that paper and the elegant results that follow all still apply. Note that one could combine these Cox process approximations with the interdomain framework and this could be a fruitful direction for further work.

\section{Conclusion and acknowledgements}

In this work we have elucidated the connection between the variational inducing point framework \citep{Titsias2009} and a rigorously defined $\KL$-divergence between stochastic processes. Early use of the rigorous formulation of $\KL$-divergence to the Gaussian processes for machine learning literature was made by Seeger \citeyearpar{Seeger2003b,Seeger2003}. Here we have increased the domain of applicability of those proofs by allowing for inducing points that are not data points, and removing unnecessary dependence on Lebesgue measure. We would argue that our proof clarifies the central and elegant role played by Radon-Nikodym derivatives. 
We then consider for the first time in this framework the case where additional variables are added solely for the purpose of variational inference. We show that marginal consistency is not enough to guarantee a principled optimization objective but that if we make the inducing points deterministic conditional on the whole function then a principled optimization objective is guaranteed and the parameters of the augmentation are variationally protected. We then show how the extended theory allows us to correctly handle principled interdomain sparse approximations and that we can cope correctly with the importance case of Cox processes where the likelihood depends on an infinite set of function points.

It seems reasonable to hope that elucidating the measure theoretic roots of the formulation will help the community to generalise the framework and lead to even better practical results. In particular it seems that since interdomain inducing points are linear functionals, the theory of Hilbert spaces might profitably be applied here. It also seems reasonable to think given the generality of section \ref{section:generalInducingPointFramework} that other Bayesian and Bayesian nonparametric models might be amenable to such a treatment.

The authors wish to thank Matthias Seeger, Michalis Titsias, Giles Shaw and the anonymous reviewer of a previous paper. AM and ZG would like to acknowledge EPSRC grant EP/I036575/1, and a Google Focused Research award. JH was supported by a MRC fellowship. RET thanks the EPSRC for funding (grant numbers EP/G050821/1 and EP/L000776/1).

\bibliography{../bib/bibliography}
\bibliographystyle{apalike} 

\end{document}